\documentclass[letterpaper, 10 pt, conference]{ieeeconf}  

\IEEEoverridecommandlockouts                              

\usepackage{amsmath} 
\usepackage{amssymb}  
\usepackage{multirow}
\usepackage{graphicx}
\usepackage{amsfonts}
\usepackage{bm}
\usepackage{multirow}
\usepackage{booktabs}
\usepackage{balance}

\title{\LARGE \bf
Learning Panorama-Aware VLA for Mobile Manipulation \\ with Whole-Body Teleoperation
}

\author{Donglin Yang$^{1,2}$, Haoran Chen$^{2,3}$, Xingyu Chen$^{4}$, Lixing Liu$^{2}$, Manyi Li$^{3}$, \\ Changhe Tu$^{3}$, Ke Xu$^{1}$, Xiaojian Ma$^{5}$\textsuperscript{\dag}, Si Liu$^{1}$\textsuperscript{\dag}
\thanks{$^\dagger$ Corresponding Authors.}
\thanks{$^{1}$Donglin Yang, Ke Xu, Si Liu are with Beihang University.    {\tt\small {yangdonglin,kexu,liusi}@buaa.edu.cn}}%
\thanks{$^{2}$Donglin yang, Haoran Chen, Lixing Liu are with Beijing Institute for General Artificial Intelligence(BIGAI). {\tt\small {yangdonglin,chenhaoran,liulixing}@bigai.ai }}%
\thanks{$^{3}$Haoran Chen, Manyi Li, Changhe Tu are with Shandong University. {\tt\small hr.chen@mail.sdu.edu.cn, {manyili,chtu}@sdu.edu.cn }}%
\thanks{$^{4}$Xingyu Chen with Johns Hopkins University.{\tt\small xchen281@jh.edu }}%
\thanks{$^{5}$Xiaojian Ma is with Delta Intelligence. {\tt\small jeasinema@gmail.com }}%
}

\begin{document}

\maketitle
\thispagestyle{empty}
\pagestyle{empty}

\begin{abstract}

Mobile manipulation is a key capability for embodied intelligence, enabling robots to accomplish complex multi-stage tasks in open-world environments. However, mobile manipulation poses two key challenges for vision-language-action (VLA) policies: At the data level, the efficient collection of high-quality whole-body demonstrations demands the coordinated control of both the mobile base and the robotic arms; at the model level, existing VLA models predominantly rely on local camera observations, whose limited field of view hinders global spatial understanding.
To address these challenges, we develop a whole-body teleoperation system and a panoramic-aware VLA policy. The system enables coordinated control of a wheeled bimanual robot through a single VR interface and supports the acquisition of a real-world mobile manipulation dataset comprising 5.5 hours of multimodal demonstrations. Building upon this dataset, we propose PanoVLA, a panorama-aware vision-language-action policy for mobile bimanual manipulation. Built upon a Mixture-of-Transformers architecture, PanoVLA introduces global spatial context through dedicated panorama encoding and fusion modules, enabling effective integration of panoramic observations with language instructions and robot states for action generation.
Evaluation on four real-world mobile manipulation tasks demonstrates that PanoVLA achieves an average stage completion rate of 91.3\% and an end-to-end success rate of 73.4\%, substantially outperforming local-view baselines. These results demonstrate that incorporating panoramic spatial context improves spatial understanding and closed-loop manipulation performance in mobile robots.

\end{abstract}

\section{INTRODUCTION}

Mobile manipulation is a central capability for general-purpose embodied agents because many real-world tasks require a robot to move through an environment, maintain spatial context, and manipulate objects with one or both arms. Recent VLA models have shown that large-scale robot data and pretrained vision-language representations can support broad manipulation skills~\cite{openx,openvla,pi05}. However, transferring this progress from tabletop manipulation to mobile bimanual robots introduces two coupled challenges: collecting coordinated whole-body demonstrations and effectively representing the surrounding environment under continuous base motion.

The first challenge is collecting demonstrations that coordinate mobility and manipulation. Recent teleoperation systems have made demonstration collection more accessible through low-cost leader--follower hardware, portable handheld interfaces, and whole-body motion tracking~\cite{aloha,gello,umi,openteach}. Other efforts have extended data collection and policy learning to mobile platforms, whole-body robots, and richer embodied sensing~\cite{mobilealoha,dass2024telemoma,momateleop,brs,robopanoptes}. Nevertheless, mobile bimanual manipulation still requires an operator to coordinate base motion, dual-arm actions, and gripper control while responding to visual feedback. When mobility and manipulation rely on separate control interfaces, demonstrations can become less fluid, especially for tasks that involve approaching, transporting, and manipulating objects across a large workspace.

The second challenge lies in achieving robust visual perception during robot motion. Conventional local observations, such as front-facing and wrist-mounted cameras, provide high-resolution manipulation cues but are inherently constrained by limited fields of view. In long-horizon mobile manipulation tasks, base translation and rotation, together with self-occlusion from the robot arms, can frequently cause target objects, landmarks, or goal regions to disappear from these local views. Existing mobile manipulation systems have therefore explored richer whole-body and multi-view sensing strategies~\cite{brs,robopanoptes}. However, simply aggregating additional image streams does not necessarily establish a coherent robot-centric global spatial understanding.
A promising alternative is to leverage panoramic imaging, which naturally provides omnidirectional observations of the surrounding environment. Nevertheless, panoramic images exhibit distinctive non-perspective distortions, creating a significant domain gap using the pretraining conventional visual encoders. As a result, directly applying standard visual representations often leads to degraded feature quality and limited preservation of semantic information.
This motivates us to develop a panorama-aware representation and fusion architecture that integrates panoramic observations with multimodal inputs and explicitly models robot-centric global context for action generation.

In this work, we investigate panorama-aware VLA learning for mobile bimanual manipulation and present PanoVLA, a panorama-enhanced VLA framework that incorporates robot-centric global context into action generation. To enable large-scale data collection, we develop a whole-body teleoperation system that tracks sparse human motion and retargets it to a wheeled bimanual robot through a General Motion Retargeting (GMR) formulation~\cite{joao2025gmr}. By modeling the mobile chassis as a planar virtual floating base, the system jointly generates base and arm commands from human demonstrations, reducing reliance on separate joystick-based base control. Using this system, we collect a real-robot dataset comprising local camera views, wrist-mounted observations, proprioceptive states, whole-body actions, and 360-degree panoramic observations from a top-mounted camera.

Based on this dataset, we propose PanoVLA, a panorama-aware three-expert Mixture-of-Transformers (MoT) VLA policy. PanoVLA integrates a panorama-specialized encoder with a dedicated panorama expert to incorporate task-conditioned, robot-centric global spatial context alongside local semantic representations for action generation.

We evaluate PanoVLA on four real-world mobile manipulation tasks involving target search, mobile approach, bimanual and single-arm manipulation, and multi-stage spatial state tracking. The collected dataset contains 800 successful teleoperated trajectories, approximately 5.5 hours of demonstrations, and 500K synchronized frames. In closed-loop real-robot evaluations, PanoVLA achieves an average stage completion rate of 91.3\% and an end-to-end success rate of 73.4\%, substantially outperforming a local-view VLA baseline with 58.6\% and 30.0\%, respectively. These results demonstrate that panoramic context provides effective global spatial reasoning for mobile manipulation.

In summary, the main contributions of this work are three-fold:
\begin{itemize}
    \item We develop a GMR-based whole-body teleoperation system that maps sparse human motion tracking to coordinated mobile-base and dual-arm commands.
    \item We collect a mobile bimanual manipulation dataset in which 360-degree panoramic observations are recorded time-synchronized with local views, wrist views, proprioceptive states, and whole-body actions, spanning 5.5 hours and 800 real-world episodes.
    \item We propose PanoVLA, a panorama-aware VLA policy that combines a panorama-specialized encoder with a panorama expert for task-conditioned spatial context fusion. Across four real-world mobile manipulation tasks, PanoVLA achieves the highest performance, with stage completion rate of 91.3\% and success rate of 73.4\%.
\end{itemize}

\section{Related Work}

\subsection{Robot Data Collection}

Expert demonstrations are crucial for robot learning, and the interface design directly affects demonstration quality, operator load, and supported embodiments. Recent teleoperation systems use head-mounted displays for intuitive control~\cite{openteach,opentelevision,bunnyvisionpro}, while portable methods capture manipulation via handheld or wearable devices~\cite{umi,dexcap,zhaxizhuoma2025fastumi}.

These methods target stationary tasks while mobile manipulation demands coordinating the motions of a mobile base and a manipulator arm. However, most existing data-collection methods decouple locomotion from manipulation at the input level and rely on dedicated base controllers such as joysticks or keyboards~\cite{brs,mobilealoha}. TeleMoMa~\cite{dass2024telemoma} supports independent control of the base and manipulator through heterogeneous input devices, including vision-based tracking and VR joysticks. MoMa-Teleop~\cite{momateleop} uses a pretrained controller to predict base velocities, requiring the operator to specify only the end-effector poses. HoMMI~\cite{xu2026hommi} instead proposes a robot-free paradigm that combines handheld grippers, wearable vision, and human-motion tracking to collect natural human motion data.
Our system uses only VR devices to track the operator's full-body motion, retargets it to the robot, and records data from the robot's onboard sensors.

\subsection{Mobile Manipulation}

General-purpose robot manipulation models~\cite{octo,openvla,rdt} provide a unified framework for perception, reasoning, and action generation. Trained on large-scale robot data, the $\pi$-series models~\cite{pi0,pi05,pi07} further improve continuous control and cross-task generalization. Beyond standard visual inputs, recent works introduce depth and tactile modalities to enrich VLA spatial and contact understanding. DepthVLA~\cite{depthvla2025} and LingBot-VLA~2.0~\cite{lingbotvla2026} infer depth geometry via learned experts or distilled tokens, while TactileAloha~\cite{tactilealoha2025} and TACT~\cite{tact2025} add tactile sensing for texture and force. This multimodal information strengthens local precision and contact robustness but is fundamentally limited to a narrow field of view.

While VLA models have shown impressive results in manipulation, extending them to mobile manipulation brings additional demands in navigation, spatial understanding, and whole-body coordination. MoManipVLA~\cite{momanipvla} proposes a hierarchical optimization framework that adapts a pretrained fixed-base manipulation VLA to mobile manipulation without large-scale retraining. End-to-end methods~\cite{yang2024harmonic, uppal2024spin} jointly learn whole-body coordination policies. SG-VLA~\cite{sgvla} adopts a lightweight VLA architecture and is jointly optimized using auxiliary spatial-grounding tasks, while AC-DiT~\cite{chen2025acditadaptivecoordinationdiffusion} learns coordinated mobile base and arm control from multimodal visual inputs. Existing methods model the environment from local views, hindering open-environment decision-making. 
To address this gap, we propose PanoVLA to exploit panoramic observations for scene-wide perception and richer spatial context.
\section{Teleoperation System and Dataset}
\label{sec:teleoperation_system}

We construct a whole-body teleoperation and data collection system around wheeled bimanual robots. Using a single wearable VR setup, the system captures sparse human motion and retargets it through GMR framework~\cite{joao2025gmr} into coordinated commands for the mobile base and dual arms. During robot execution, the system synchronously records local RGB views, a top-mounted 360-degree panoramic observation, proprioceptive states, and whole-body actions. This section describes the teleoperation hardware, the GMR-based retargeting procedure, and the data collection protocol.

\subsection{Teleoperation System}
\label{sec:sys_overview}

\begin{figure}[t]
\centering
\includegraphics[width=0.98\linewidth]{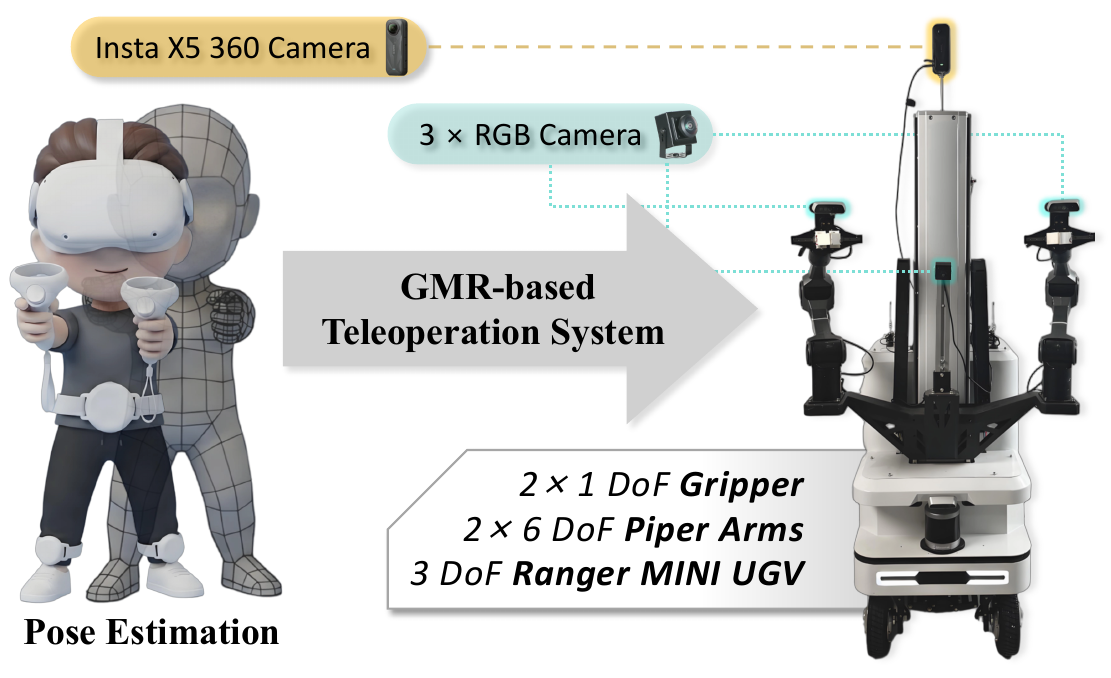}

\caption{Overview of the whole-body teleoperation pipeline. Sparse human-motion measurements acquired with the wearable VR interface are reconstructed into an SMPL model and retargeted through GMR to the wheeled bimanual robot. }
\label{fig:teleop_pipeline}
\vspace{-3mm}
\end{figure}

An overview of the teleoperation pipeline is shown in Figure~\ref{fig:teleop_pipeline}, which consists of VR-based human motion capture and pose estimation, SMPL reconstruction, GMR retargeting, and the robot hardware.

\medskip
\noindent\textbf{Operator interface.} The operator wears a VR headset for 6-DoF head tracking, holds one controller in each hand to track hand poses and issue gripper commands, and wears three body trackers: one at the waist and one at each ankle. Similar to recent wearable and mixed-reality teleoperation systems~\cite{openteach,opentelevision,bunnyvisionpro}, this sparse tracking interface is portable and easy to deploy, and does not require an input device tailored to a specific robot embodiment. The system reconstructs the sparse tracking stream into an SMPL body state in real time, from which pelvis motion, foot motion, and bilateral hand poses are extracted as retargeting inputs. During a demonstration, the operator may either share the physical workspace with the robot and directly observe its execution, or teleoperate from a separate physical space using real-time video streamed from the onboard cameras to the VR headset.

\begin{figure*}[!t] 
\centering
\includegraphics[width=0.99\linewidth]{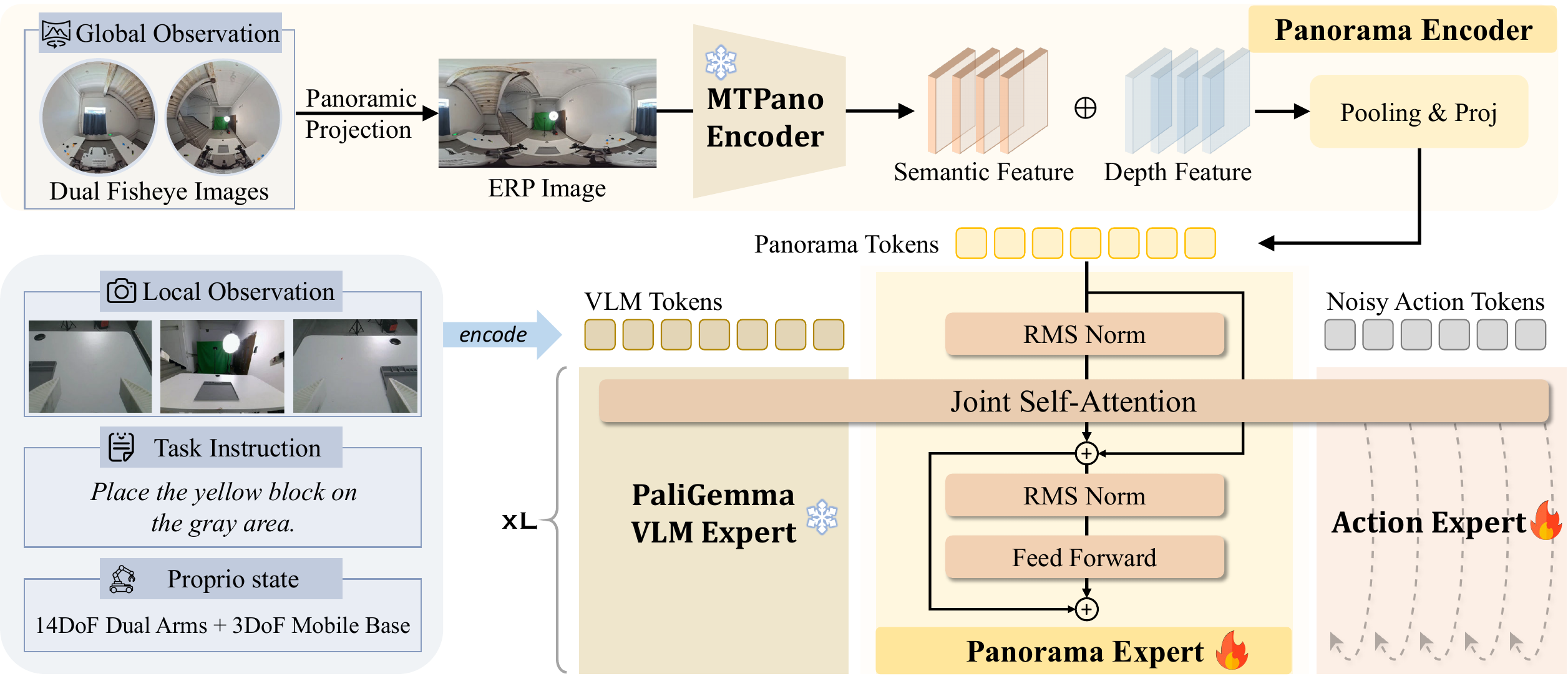}
\caption{Overview of the proposed PanoVLA framework, which incorporates panoramic context into a vision-language-action model for mobile manipulation.}
\label{fig:framework-panovla}
\vspace{-3mm}
\end{figure*}

\medskip
\noindent\textbf{Robot embodiment.} As shown on the right side of Fig.~\ref{fig:teleop_pipeline}, the robot is a wheeled bimanual platform comprising a RANGER MINI 3 mobile base, two 6-DoF Agilex PIPER arms, and parallel grippers. Its visual sensing suite consists of a top-mounted 360-degree panoramic camera and three local RGB cameras: one front-facing camera and two wrist-mounted cameras. The mobile base provides planar mobility and supports spin, ackermann-steering, and diagonal-translation modes.

\medskip
\noindent\textbf{Motion Retargeting.}
The retargeting module takes human motion features reconstructed from SMPL as input and maps them to robot whole-body commands through GMR optimization. These features include the pelvis pose, bilateral hand positions and orientations, and lower-body motion cues. We parameterize the robot configuration as
\(q = (x_b, y_b, \theta_b, q_L, q_R),\)
where \((x_b,y_b,\theta_b)\) denotes the planar base pose in the local teleoperation frame, and \(q_L\) and \(q_R\) denote the left- and right-arm joint configurations, respectively. Gripper commands are generated from the controller trigger states and recorded with the same timestamp as the optimized base and arm commands.

At each control step, the robot configuration is obtained by minimizing a weighted objective:
\begin{equation}
\begin{aligned}
q^*=\arg\min_q \;&
\lambda_{pos}\mathcal{L}_{pos}
+ \lambda_{rot}\mathcal{L}_{rot} \\
&+ \lambda_{base}\mathcal{L}_{base}
+ \lambda_{reg}\mathcal{L}_{reg}.
\end{aligned}
\label{eq:gmr_objective}
\end{equation}
After the human motion is expressed in the robot teleoperation frame, \(\mathcal{L}_{pos}\) aligns the hand and gripper positions, while \(\mathcal{L}_{rot}\) aligns the corresponding hand and end-effector orientations. \(\mathcal{L}_{base}\) couples planar base motion to the operator's pelvis displacement and facing direction, and \(\mathcal{L}_{reg}\) discourages joint-limit violations and excessive configuration changes. This objective follows the design principles of whole-body human-to-robot retargeting systems~\cite{h2o,omnih2o,twist}, while adapting the floating-base motion constraints to a wheeled bimanual platform.

\medskip
\noindent\textbf{Mobile manipulation adaptation.} In the retargeting optimization, we model the mobile base as a 3-DoF planar virtual floating base and optimize its pose jointly with the dual-arm joint configurations. Because the base pose participates in the end-effector alignment objective, the optimizer can adjust the base translation and yaw when a mapped target end-effector pose lies outside the current arm workspace, thereby reducing the retargeting error and the tendency of the arm joints to approach their limits. The optimized base motion is converted into a velocity command \((v_x,v_y,\omega)\). This joint optimization allows the operator to coordinate the mobile base and dual arms through natural body motion. 

To execute the virtual 3-DoF commands on the physical base, we propose a cost-driven mode-switching strategy. At each control cycle, the strategy evaluates the matching cost between three motion modes and the desired translation and rotation, and selects the mode with the lowest cost for execution. After the selected mode executes for a short time step, the strategy recomputes the cost based on the latest retargeting output to decide whether to keep or switch to another. This periodic decision mechanism enables the base motion to dynamically follow changes in the operator’s intent. At each control step, the system synchronously sends the base-velocity command, dual-arm joint-position targets, and gripper commands, and produces a time-aligned whole-body action sequence.

\subsection{Data Collection Protocol and Dataset}
\label{sec:dataset}

Using the teleoperation system above, we collect 200 demonstration trajectories for each of the four real-world mobile manipulation tasks, yielding 800 trajectories in total. During each demonstration, the system synchronously records visual observations from all cameras, robot proprioceptive states, and whole-body actions with shared timestamps. The trajectories span approximately 5.5 hours in total. Each trajectory lasts about 20--30 seconds and is recorded at 25~Hz, yielding roughly 500k synchronized frames.

\medskip
\noindent\textbf{Data Modalities and Representation}
Figure~\ref{fig:teleop_pipeline} shows the visual sensors and viewpoints recorded in the dataset. At each time step, a data record contains the local visual observation, panoramic observation, proprioceptive state, and whole-body action. The local observation comprises RGB images captured by the front-facing camera and two wrist-mounted cameras. These views cover the scene in front of the robot and the manipulation regions around the end effectors, providing local visual cues for target approach and fine-grained manipulation. The panoramic observation is captured by the top-mounted 360-degree camera and converted into and stored as an equirectangular projection (ERP) image. Compared with the local views, it covers a broader spatial extent around the robot and provides robot-centric context for target search, mobile approach, and multi-stage task execution.

The proprioceptive state includes the absolute joint positions of the two 6-DoF arms, the gripper open/close states, and the planar base pose \((x,y,\theta)\) relative to the robot frame at the start of the trajectory. The whole-body action \(a_t\in\mathbb{R}^{17}\) contains 14 dimensions for the dual-arm joints and grippers, together with the three-dimensional base command comprising the planar velocity \((v_x,v_y)\) and yaw rate \(\omega\).

\begin{table}[t]
\centering
\caption{Task taxonomy for the panorama-aware mobile manipulation dataset, summarizing the task objectives, target layouts, and spatial settings.}
\label{tab:task_taxonomy}
\small
\resizebox{0.99\columnwidth}{!}{%
\begin{tabular}{@{}lll@{}}
\toprule
Task & Goal & Spatial setting \\
\midrule
Move Pen & place the pen in its holder & front / left / right \\
Move Block & transfer blocks to the front area & left / right / both sides \\
Open Curtain & open the curtains on both sides & left / right \\
Wipe Table & wipe stains and return the cloth & along a long table \\
\bottomrule
\end{tabular}
}
\vspace{-4mm}
\end{table}

\medskip
\noindent\textbf{Task Taxonomy}
\label{sec:task_taxonomy}
To cover real-world mobile manipulation across diverse spatial layouts and workspace extents, we define four tasks. Table~\ref{tab:task_taxonomy} summarizes the objective and spatial setting of the tasks in our dataset. The task suite varies target placement, bilateral interaction, and spatial extent: \emph{Move Pen} and \emph{Move Block} change the target direction relative to the robot, \emph{Open Curtain} requires sequential interaction with both sides, and \emph{Wipe Table} extends manipulation over a long table before returning the cloth. In these tasks, task-relevant regions are not simultaneously covered by the local cameras, creating a need for cross-view spatial reasoning and multi-stage state tracking.



\section{Panorama-Enhanced Mobile Manipulation}
\label{sec:pano_method}

\subsection{Problem Formulation}
\label{subsec:pano_formulation}
To provide global spatial context for language-conditioned mobile manipulation, we augment the robot observation with a robot-centric panoramic view. Specifically, the observation is defined as
\(
o_t=\{I_t^{\mathrm{loc}},I_t^{\mathrm{pano}},s_t\},
\)
where $I_t^{\mathrm{loc}}$ denotes local visual observations, $I_t^{\mathrm{pano}}$ denotes the panoramic observation covering the robot's surrounding environment, and $s_t$ is the proprioceptive state.

Given the observation $o_t$ and a language instruction $l$, the policy $\pi_\theta$ predicts a horizon-$H$ continuous action sequence,
\begin{equation}
\hat a_{t:t+H-1}=\pi_\theta(o_t,l),
\end{equation}
where each $a_t$ is a continuous whole-body action comprising commands for the mobile base, dual arms, and grippers.
\subsection{PanoVLA}
\vspace{-2mm}
\label{subsec:panovla}

\medskip
\noindent\textbf{Model Architecture.}
PanoVLA is a panorama-aware Mixture-of-Transformers VLA policy with three expert blocks. The VLM expert encodes the local views, instruction, and robot state; the panorama expert converts the wide-field observation into task-conditioned spatial context; and the action expert predicts continuous action chunks. We employ a block-wise causal attention mask, tokens attend bidirectionally within their own context block, while a later block can attend only to its own and preceding blocks. PanoVLA is instantiated from the dual-expert \(\pi_{0.5}\) backbone by inserting the panorama expert between its VLM and action experts.

We adopt the transformer key--value (KV) cache as the fusion interface. For an $L$-layer expert, a cache is the layer-wise collection
\(
C=\{(K^{\ell},V^{\ell})\}_{\ell=1}^{L}
\)
generated from its context tokens and reused by subsequent tokens during attention. At time $t$, the VLM expert computes the VLM cache $C_t^{\mathrm{vlm}}$ from $I_t^{\mathrm{loc}}$, $l$, and $s_t$. In parallel, the panorama encoder encodes $I_t^{\mathrm{pano}}$ into tokens $Z_t^{\mathrm{pano}}$. The panorama expert then performs joint self-attention with $C_t^{\mathrm{vlm}}$ while processing $Z_t^{\mathrm{pano}}$, and produces panorama cache $C_t^{\mathrm{pano}}$. The action expert subsequently predicts the action chunk based on the combined cache \(C_t=[C_t^{\mathrm{vlm}};C_t^{\mathrm{pano}}]\), where $[\cdot;\cdot]$ denotes concatenation along the cached-token dimension.
This design enables the action expert to jointly leverage fine-grained manipulation cues from local observations and robot-centric global context from panoramic observations.




\medskip
\noindent\textbf{Panorama Encoder.}
The panorama encoder processes panoramic observations into a unified dense representation that captures both semantic and geometric information, which is subsequently adapted into compact visual tokens for the VLA backbone. As illustrated in Fig.~\ref{fig:framework-panovla}, the pipeline consists of three stages: spherical projection, panoramic feature extraction, and token adaptation. Given synchronized dual-fisheye observations, we first apply spherical reprojection to map the two fisheye views onto a common viewing sphere and rasterize the spherical representation as an equirectangular projection (ERP) image, yielding a continuous, robot-centric panorama of the surroundings. This ERP image is then fed into a panoramic visual encoder $E_{\mathrm p}$. 
We instantiate \(E_{\mathrm{p}}\) with MTPano~\cite{zhang2026mtpanomultitaskpanoramicscene}, a panoramic foundation model for dense scene parsing. Specifically, we extract intermediate features from its segmentation and depth branches, obtaining two dense features that capture semantic and geometric cues.
These are then concatenated along the channel dimension to form a unified dense panoramic representation.

To enable panoramic observations to be integrated into multimodal understanding, the fused feature map is further transformed into the token embedding space of the VLA backbone through a lightweight token adapter $A_{\psi}$. Specifically, adaptive average pooling first compresses this dense representation into a fixed spatial resolution, producing a constant number of visual tokens independent of the panorama resolution, a lightweight multi-layer perceptron (MLP) layers then map each token into the hidden dimension of the panorama expert, yielding the panorama token sequence.
\begin{equation}
Z_t^{\mathrm{pano}}
=
A_{\psi}
\left(
E_{\mathrm{p}}
\left(
\Pi_{\mathrm{sph}}
\left(
I_t^{\mathrm{pano}}
\right)
\right)
\right),
\end{equation}
where \(\Pi_{\mathrm{sph}}\) denotes the spherical reprojection operator, and $Z_t^{\mathrm{pano}}\in\mathbb{R}^{N_p\times d}$ represents the visual tokens of the panorama. Here, $N_p$ is the number of panoramic tokens, and $d$ is the hidden dimension of the panorama expert. These panorama tokens serve as input to the panorama expert for subsequent global spatial understanding.

\medskip
\noindent\textbf{Panorama Expert.}
The panorama expert performs task-conditioned spatial encoding by integrating panoramic observations with the semantic context provided by the VLM expert. Given panorama tokens \(Z_t^{\mathrm{pano}}\) and the VLM cache \(C_t^{\mathrm{vlm}}\), it produces a panoramic cache via joint self-attention, capturing task-relevant global spatial information for downstream action prediction:
\begin{equation}
C_t^{\mathrm{pano}} = \phi_{\mathrm{pano}}(Z_t^{\mathrm{pano}} \mid C_t^{\mathrm{vlm}}, M_{\mathrm{pano}}),
\end{equation}
where the panoramic cache is computed under a block-wise causal mask \(M_{\mathrm{pano}}\).

\medskip
\noindent\textbf{Training Objective.}
Conditioned on the joint context \(C_t\), which integrates both the semantic context from the VLM expert and the panoramic spatial context from the panorama expert, the action expert predicts continuous action chunks using conditional flow matching. Specifically, for a demonstrated action chunk \(a\), a noisy action sample \(x_{\tau}\), and a flow-matching timestep \(\tau\), the model learns a conditional velocity field by minimizing
\begin{equation}
\mathcal{L}=
\mathbb{E}_{p(a\mid o_t,l),\,q(x_{\tau}\mid a)}
\left[
\|v_{\theta}^{\mathrm{act}}(x_{\tau},\tau\mid C_t)-u(x_{\tau}\mid a)\|_2^2
\right],
\label{eq:fm_objective}
\end{equation}
where \(x_{\tau}\) denotes a noisy action sample along the flow trajectory, \(u(x_{\tau}\mid a)\) is the target velocity field, and \(v_{\theta}^{\mathrm{act}}\) is the velocity field predicted by the action expert.

\section{Experiments}
\label{sec:experiments}

\begin{figure*}[t] 
\centering
\includegraphics[width=1.0\linewidth]{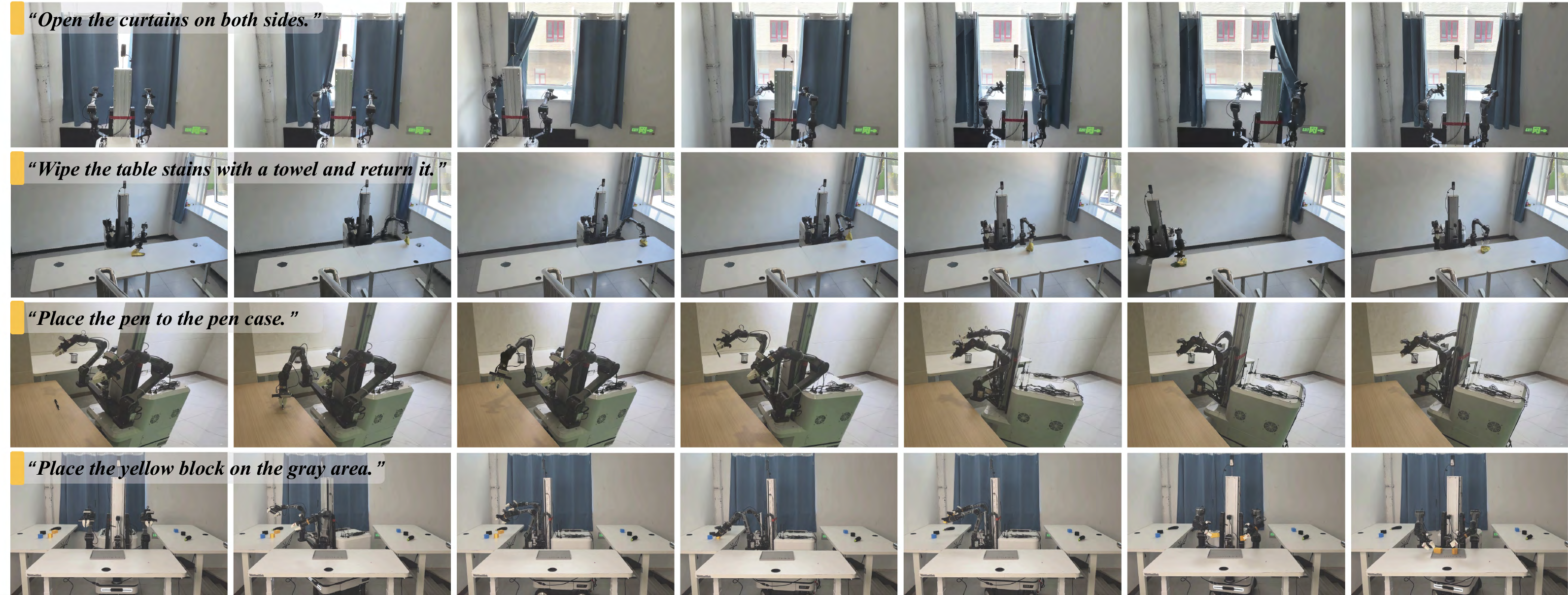}
\caption{Representative real-world closed-loop rollouts across the four mobile manipulation tasks. Across multiple large-workspace, multi-stage tasks, PanoVLA leverages panoramic context for spatial understanding and task-state tracking, while effectively coordinating the mobile base and dual arms.
}
\label{fig:real-world-exp}
\end{figure*}

\begin{table*}[!t]
\centering
\caption{Real-world closed-loop evaluation. SCR denotes stage completion rate and SR denotes end-to-end success rate; Avg. is averaged over the four tasks, with 15 trials per task. The best results are highlighted in \textbf{bold}.}
\label{tab:real_world_results}
\renewcommand{\arraystretch}{0.92}
\small
\resizebox{0.98\textwidth}{!}{%
\begin{tabular}{l|cc|cc|cc|cc|cc}
\toprule
\multirow{2}{*}{Method} &
\multicolumn{2}{c|}{Move Pen} &
\multicolumn{2}{c|}{Move Block} &
\multicolumn{2}{c|}{Open Curtain} &
\multicolumn{2}{c|}{Wipe Table} &
\multicolumn{2}{c}{Avg.} \\[-0.6pt]
\cmidrule(lr){2-11}
& SCR & SR & SCR & SR & SCR & SR & SCR & SR & SCR & SR \\[-0.8pt]
\midrule
\(\pi_{0.5}\) & 71.1\% & 46.7\% & 35.0\% & 20.0\% & 61.7\% & 26.7\% & 66.7\% & 26.7\% & 58.6\% & 30.0\% \\
\(\pi_{0.5}\) w/ Pano & \textbf{95.6\%} & \textbf{86.7\%} & 80.0\% & 73.3\% & 73.3\% & 46.7\% & 74.4\% & 20.0\% & 80.8\% & 56.7\% \\
\(\pi_{0.5}\) w/ Stacked Pano & 82.2\% & 60.0\% & 70.0\% & 40.0\% & 56.7\% & 13.3\% & 72.8\% & 40.0\% & 70.4\% & 38.3\% \\
\midrule
\textbf{PanoVLA} & \textbf{95.6\%} & \textbf{86.7\%} & \textbf{96.7\%} & \textbf{93.3\%} & \textbf{88.3\%} & \textbf{66.7\%} & \textbf{84.4\%} & \textbf{46.7\%} & \textbf{91.3\%} & \textbf{73.4\%} \\
\bottomrule
\end{tabular}
}
\vspace{-4mm}
\end{table*}

\subsection{Experimental Setup}
\label{subsec:exp_setup}

\noindent\textbf{Baselines.}
To isolate the effect of panorama modeling, we compare PanoVLA with three controlled \(\pi_{0.5}\)-based baselines that vary only in how panoramic observations are represented and incorporated into the policy.

\begin{itemize}
    \item \textbf{\(\bm{\pi}_{\mathbf{0.5}}\).} The policy encodes local camera views using SigLIP and takes language instructions together with robot proprioceptive states as input.

    \item \textbf{\(\bm{\pi}_\mathbf{0.5}\) w/ Pano.} The equirectangular panorama is treated as an additional image input and encoded by the same SigLIP encoder as the local camera views. No dedicated panorama branch is introduced.

    \item \textbf{\(\bm{\pi}_\mathbf{0.5}\) w/ Stacked Pano.} The panorama is first reprojected into three perspective views whose optical axes are uniformly spaced by \(120^\circ\) in yaw. The three views are concatenated into a single composite image, which is then provided as an additional visual input alongside other local images to the original \(\pi_{0.5}\) architecture, following prior vision-language methods for robot learning~\cite{duan2024aha,song2026rethinking} that use stitched visual inputs.

    \item \textbf{PanoVLA.} Our model encodes panoramic observations with MTPano and fuses panoramic features through the panorama expert before action generation.
\end{itemize}

\noindent\textbf{Evaluation Protocol and Metrics.}
We evaluate each method on the four tasks introduced in Section~\ref{sec:task_taxonomy}.
To enable fine-grained failure analysis, we decompose each task into multiple stages and evaluate each method with 15 closed-loop real-robot trials per task. Every trial is conducted as a single uninterrupted rollout, with all actions generated by the policy and no manual resets of failed stages. Under this protocol, we report two metrics: success rate (SR) and stage completion rate (SCR). SR is a binary trial-level metric that records whether the full task is completed, whereas SCR averages the proportion of applicable stages completed across all trials.



\noindent\textbf{Implementation Details.}
We build PanoVLA upon the \(\pi_{0.5}\) VLA framework, using Gemma-2B as the language backbone and SigLIP for local camera observations. PanoVLA augments the policy with a pretrained MTPano encoder and a 100M-parameter panorama expert for panoramic representation, while retaining the action expert for action generation. For all methods, both the proprioceptive state and action are represented as 17-dimensional vectors, normalized using the 1st and 99th percentiles. At each time, the policy predicts a 32-step action chunk for the arm and base, expressed relative to the current proprioceptive state; gripper actions are represented separately as binary open/close commands. During training, we freeze the language model and the pretrained MTPano encoder and optimize the remaining modules.

\noindent\textbf{Training Configuration.}
We independently fine-tune a policy using the corresponding task-specific demonstrations from the dataset described in Section~\ref{sec:dataset}. Within each task, all methods use the same training demonstrations and optimization settings. We optimize all trainable parameters using AdamW with the gradient norm clipped to 1.0. The model is trained for 30K optimization steps. The learning rate is linearly warmed up over the first 5K steps and fixed at 5e-5 thereafter. Training is performed in bfloat16 precision on four NVIDIA A100 GPUs with a per-GPU batch size of 64. During real-robot evaluation, the policy runs in closed loop on a single NVIDIA RTX 4090 GPU.

\subsection{Real-World Experiments}
\label{subsec:real_world_exp}

Table~\ref{tab:real_world_results} summarizes the closed-loop real-robot performance across the four tasks.
PanoVLA achieves the best average results, with an SCR of \(91.3\%\) and an SR of \(73.4\%\); the standard \(\pi_{0.5}\) baseline attains \(58.6\%\) SCR and \(30.0\%\) SR, respectively.
The substantial gain in SCR indicates that PanoVLA performs better across intermediate stages such as target search, mobile approach, and physical manipulation.

Mobile manipulation often requires robots to localize targets beyond the current field of view and navigate across a larger workspace to reach them, making local-view policies prone to failures.
The standard \(\pi_{0.5}\) policy depends exclusively on local camera views and frequently fails to infer the spatial location of the target object or manipulation area relative to the robot. Although panoramic observations provide additional spatial context and generally improve performance, the inconsistent gains among panorama-based variants suggest that effective panoramic perception relies not merely on broader views, but on how panoramic geometry is represented and incorporated into the policy.

\begin{figure}[t] 
\centering
\includegraphics[width=1.0\linewidth]{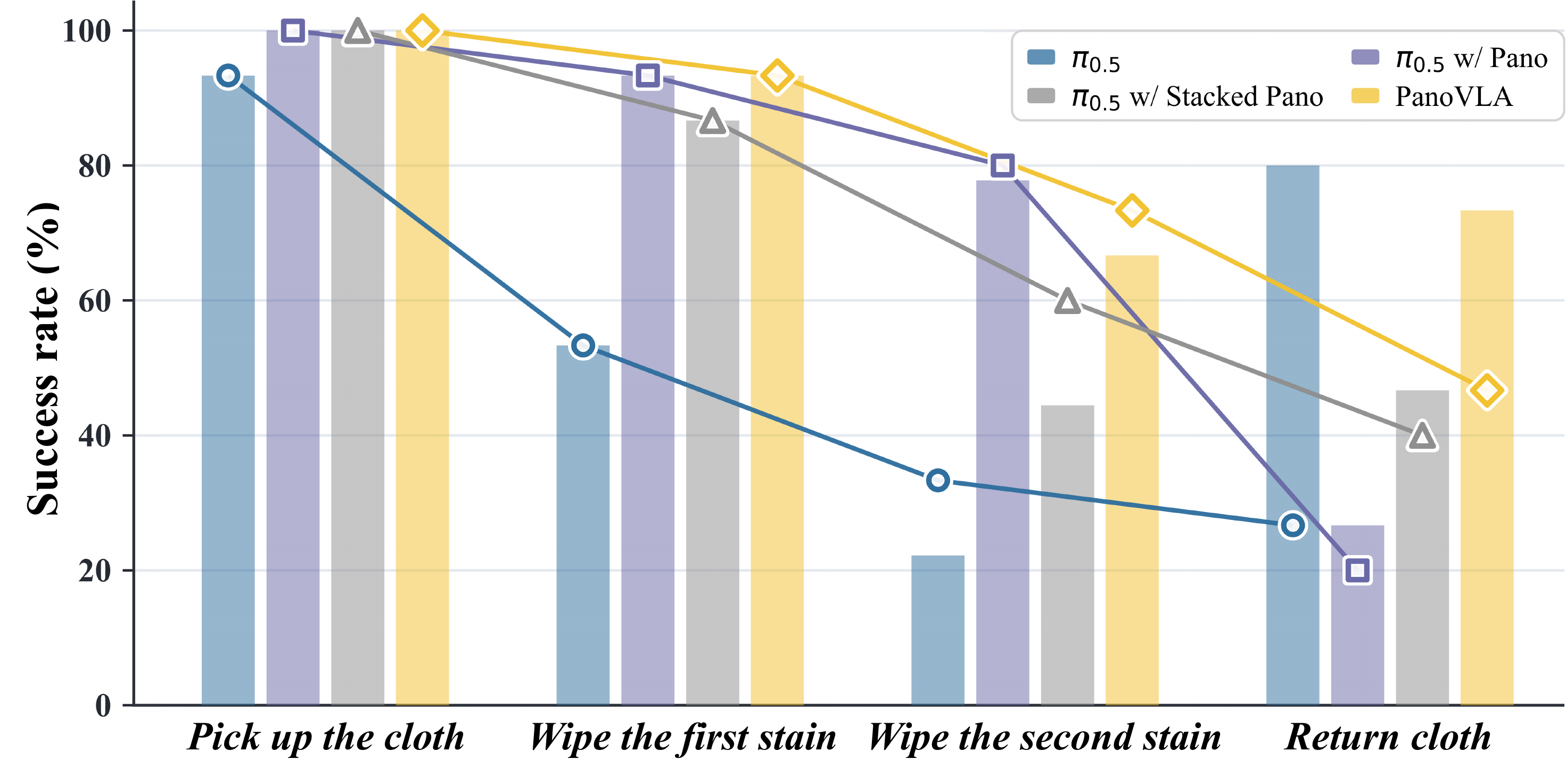}

\caption{Stage-wise performance on Wipe Table task. Bars summarize per-stage completion performance, while dashed lines trace cumulative success.}
\label{fig:wipe-table-progress-sr}
\vspace{-5mm}
\end{figure}
\(\pi_{0.5}\) w/ Pano matches PanoVLA on \emph{Move Pen}
, as this task mainly involves single‑target localization and short‑horizon pick‑and‑place. The equirectangular panorama already provides sufficient spatial information to \(\pi_{0.5}\), reducing the benefit of explicitly modeling panoramic information with panorama expert. The PanoVLA advantage becomes more apparent in tasks that require multistage state tracking ‐ and long-range spatial consistency.
Although the raw panorama input baseline performs well on \emph{Move Pen}, its \emph{Wipe Table} profile in Fig.~\ref{fig:wipe-table-progress-sr} drops at the final stage.
It completes this stage in only \(26.7\%\) of trials, resulting in an SCR of \(74.4\%\) but an end-to-end SR of only \(20.0\%\), whereas PanoVLA achieves \(73.3\%\) stage completion and \(46.7\%\) SR. This substantial gap suggests that the panorama expert more effectively tracks task state transitions while preserving global spatial consistency across multiple manipulation stages, enabling the robot to reliably relocalize the cloth and return it to its original position.

Reprojecting the panorama into perspective views introduces a different trade-off.
The stacked-panorama baseline partially alleviates the mismatch between equirectangular panoramas and perspective-image encoders by decomposing the panorama into multiple perspective views.
On \emph{Wipe Table}, its observed SR is \(40.0\%\), compared with \(20.0\%\) for raw panorama input, indicating that perspective reprojection can benefit manipulation-heavy stages.
However, it underperforms the raw panorama baseline on the remaining tasks, suggesting that perspective crops may weaken the cross-view continuity that benefits robot-centric understanding.

The task-wise comparison further suggests that difficulty is not solely determined by whether the target lies within the current field of view. \emph{Move Pen} and \emph{Move Block} primarily depend on target localization and succeed reliably once the target direction is correctly estimated, whereas \emph{Open Curtain} and \emph{Wipe Table} demand continuous state tracking and spatial understanding across stages. Despite these differing requirements, PanoVLA achieves the highest SR on all four tasks. This consistent advantage indicates that the robot-centric panoramic representation not only improves spatial awareness but also stabilizes closed-loop policy execution over long horizons.

\subsection{Ablation Studies}
\label{subsec:ablation}

\begin{table}[!t]
\centering
\caption{Effect of expert size and panorama encoder selection.}
\label{tab:ablation_panorama}
\scriptsize
\renewcommand{\arraystretch}{0.95}
\resizebox{0.98\columnwidth}{!}{%
\begin{tabular}{l|c|cc}
\toprule
Variant & Expert Size & SCR & SR \\
\midrule
\multicolumn{4}{l}{\textit{Expert size}} \\[-0.6pt]
\cmidrule(lr){1-4}
PanoVLA w/ MTPano & 50M & 84.4\% & 60.0\% \\
PanoVLA w/ MTPano & 100M & \textbf{95.6\%} & \textbf{86.7\%} \\
PanoVLA w/ MTPano & 200M & 93.3\% & \textbf{86.7\%} \\
PanoVLA w/ MTPano & 300M & 91.1\% & 73.3\% \\
\midrule
\multicolumn{4}{l}{\textit{Panorama encoder}} \\[-0.6pt]
\cmidrule(lr){1-4}
PanoVLA w/ SigLIP & 100M & 75.6\% & 40.0\% \\
PanoVLA w/ MTPano & 100M& \textbf{95.6\%} & \textbf{86.7\%} \\
\bottomrule
\end{tabular}
}
\vspace{-4mm}
\end{table}

We conduct all ablation studies on the \emph{Move Pen} task, evaluating each variant over 15 closed-loop real-robot trials.

\noindent\textbf{Panorama expert scale.}
We investigate the impact of panorama expert capacity by varying the parameter count as shown in Table~\ref{tab:ablation_panorama}. These results suggest that the 50M expert lacks sufficient capacity to capture cross-view dependencies and target-direction cues, while scaling beyond 100M parameters does not lead to consistent closed-loop improvements given the limited fine-tuning data. We therefore adopt the 100M expert, providing a favorable empirical trade-off between representational capacity and data-efficient fine-tuning effectiveness.

\noindent\textbf{Panorama visual encoder.}
We further investigate the impact of panorama encoder choice by comparing SigLIP, the vision encoder of the base VLM, with MTPano.
Using SigLIP yields only \(75.6\%\) SCR and \(40.0\%\) SR, substantially lower than the \(95.6\%\) and \(86.7\%\) achieved with MTPano.
Because SigLIP is pretrained primarily on perspective images, it may be less effective at capturing the horizontal continuity and global layout of equirectangular observations.
In contrast, MTPano is designed to model panorama-specific spatial structure, providing the panorama expert with more informative cues for constructing robot-centric spatial context.
These results support our design choice of using geometry-aware panoramic representations to construct spatial context.
\section{Conclusion}
\label{sec:conclusion}
This paper proposes a unified framework for panorama-aware mobile manipulation learning, encompassing a whole-body data collection system for mobile manipulation and a closed-loop policy learning method. Within a general motion retargeting pipeline, the teleoperation system enables operator to coordinate control of the mobile base, dual robotic arms, and grippers through natural body movements, without requiring an additional motion controller. We collected a 5.5 hours demonstration dataset of 800 episodes, including panoramic and local observations, proprioceptive states and whole-body actions. Building on this dataset, we propose PanoVLA, an MOT-based architecture that introduces a dedicated panoramic encoder and a panoramic expert module to inject task-relevant panoramic information into the action expert. In four real-world mobile manipulation tasks with diverse spatial layouts, PanoVLA outperforms the baseline model using local views in both stage completion rate and end-to-end success rate, providing a robust and scalable solution for panorama-aware mobile manipulation learning.

\newpage
\balance
\bibliographystyle{IEEEtran}
\bibliography{IEEEexample} 

\end{document}